\documentclass[letterpaper]{article} 
\usepackage[preprint]{aaai2027}
\usepackage[hyphens]{url}  
\usepackage{graphicx} 
\usepackage{natbib}  
\usepackage{caption} 
\usepackage{tabularx}
\usepackage{amsfonts}
\usepackage{amsmath}
\usepackage{algorithm}
\usepackage{algorithmic}

\usepackage{newfloat}
\usepackage{listings}
\DeclareCaptionStyle{ruled}{labelfont=normalfont,labelsep=colon,strut=off} 
\floatstyle{ruled}
\newfloat{listing}{tb}{lst}{}
\floatname{listing}{Listing}
\usepackage{pifont}
\newcommand{\cmark}{\ding{51}}
\newcommand{\xmark}{\ding{55}}

\graphicspath{{../}{./}}
\usepackage{booktabs}   
\usepackage{makecell}   
\usepackage{placeins}   
\usepackage{multirow}  
\title{RenoLayout: Unified Floorplan Renovation and Furniture Layout for Constraint-Aware Residential Design}
\author{
    Jingguo Liu,
    Yuanqing Liu,
    Chong Zhang,
    Yue Yang\corresponding
}
\affiliations{
    Beike\\
    \texttt{\{liujingguo002, liuyuanqing006, zhangchong052, yangyue092\}@ke.com}
}

\begin{document}

\maketitle

\begin{abstract}
Residential design aims to transform existing homes into functional living spaces while satisfying diverse design preferences and physical constraints. However, existing floorplan generation methods typically assume empty building boundaries as input, whereas renovation requires reasoning over existing floorplans. Furthermore, floorplan renovation and furniture layout are commonly addressed as separate tasks, overlooking their inherent spatial dependency. Consequently, practical residential design must preserve immutable structural components while transforming editable spatial elements. It must also optimize furniture placement under design objectives and physical constraints. To address these challenges, we formulate residential design as a design-point-guided conditional generation task and propose \textbf{RenoLayout}, a framework that translates a pre-design residential state together with a design-point set into an executable \textit{Geometry-as-Code} program. To enable controllable supervision, we construct a \textbf{Design Map} comprising 123 fine-grained design points covering structural renovation and furniture layout. We further introduce \textbf{Difference-Grounded Design-Point Annotation} to derive design-point sets automatically from before--after design pairs with VLM-based consistency verification. RenoLayout is trained using a three-stage CPT--SFT--RL pipeline, progressively improving spatial reasoning, instruction following, and constraint compliance. We also introduce the \textbf{RenoLayout Benchmark}, a set of 200 real-world before--after renovation cases with complete furniture layouts. To the best of our knowledge, it is the first benchmark for unified residential design. These cases include design-critical structural and building-service constraints. Extensive experiments evaluate design-point fulfillment, constraint compliance, living quality, and professional approval. On the RenoLayout Benchmark, RenoLayout achieves 91.63\% Recall@1 and 61.00\% Pass@1. These results demonstrate effective design-point fulfillment while maintaining constraint compliance.
\end{abstract}


\section{Introduction}

Residential design transforms an existing home into a coordinated spatial solution that satisfies user requirements while respecting physical constraints. Unlike generation within an empty boundary, this task must preserve the exterior envelope, load-bearing walls, shafts, pipes, and drainage points. Meanwhile, editable partitions, room functions, and furniture layouts may be jointly modified.

\begin{figure}[t]
    \centering
    \includegraphics[width=0.95\linewidth]{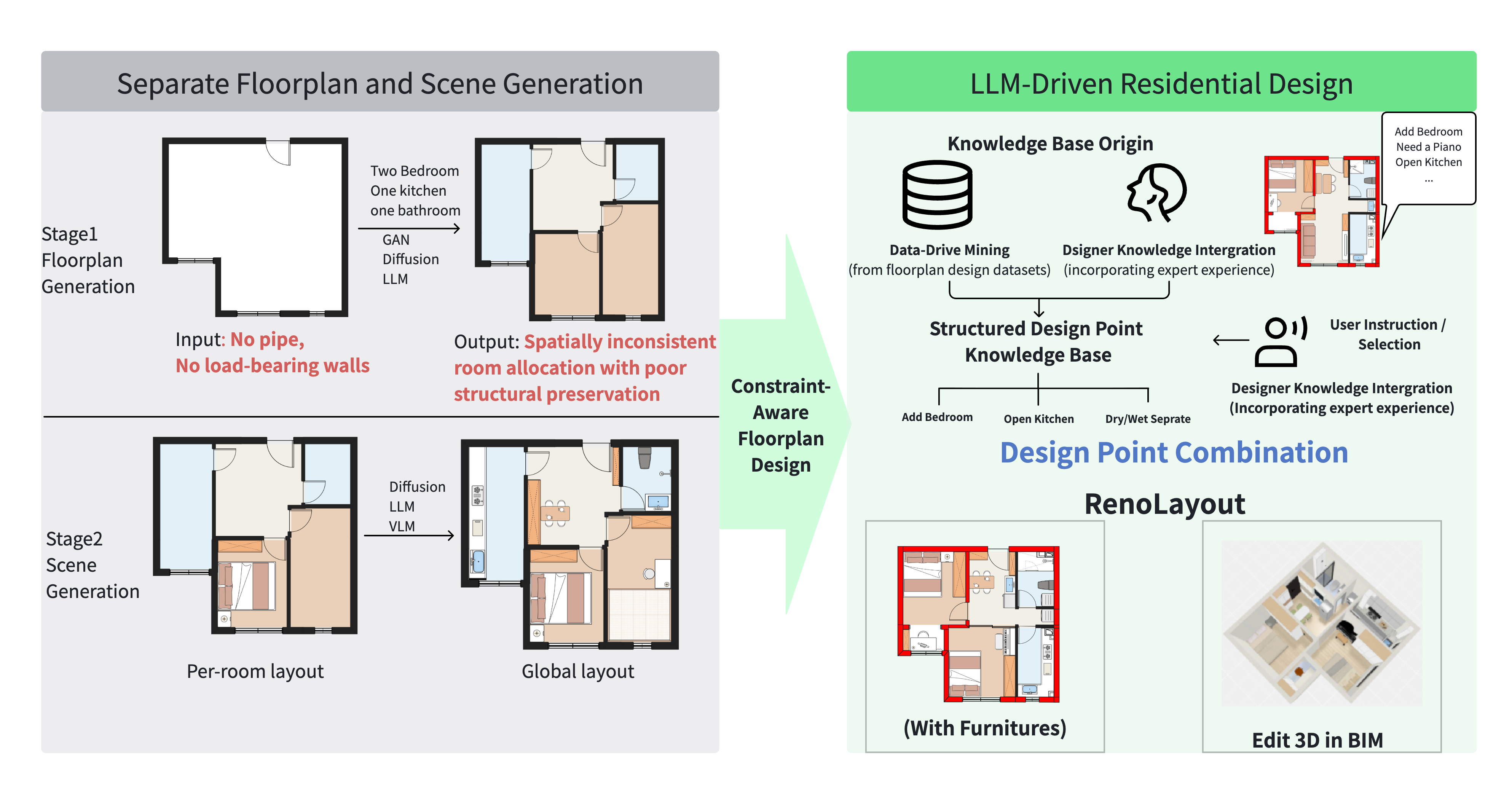}
    \caption{Comparison of existing task-specific paradigms with RenoLayout. Existing methods treat floorplan generation separately from furniture layout, whereas RenoLayout unifies constraint-aware renovation with full-home furnishing.}
    \label{fig:compare_fu}
\end{figure}
As illustrated in Figure~\ref{fig:compare_fu}, floorplan methods focus on room partitioning and structural planning \cite{House-gan, yin2025floorplan, housemind}, whereas furniture-layout methods target furniture arrangement with functional composition \cite{feng2023layoutgpt, yang2025optiscene, hoglayout, lavigen}. Joint approaches such as HouseDiffusion and Nestwork generate furnished layouts \cite{shabani2023housediffusion, nestwork}, but their data generally describe only final designs rather than paired before--after cases with complete furnishing under design-critical constraints. Table~\ref{tab:data_annotation_comparison} summarizes this supervision gap.
\begin{table*}[t]
\centering
\scriptsize
\setlength{\tabcolsep}{2.5pt}
\renewcommand{\arraystretch}{1.05}
\begin{tabularx}{0.95\textwidth}{>{\raggedright\arraybackslash}X *{6}{>{\centering\arraybackslash}p{1.75cm}}}
\toprule
\makecell[l]{\textbf{Task setting}\\\textbf{Representative works}}
& \makecell[c]{\textbf{Pre-design}\\\textbf{state}}
& \makecell[c]{\textbf{Post-design}\\\textbf{solution}}
& \makecell[c]{\textbf{Full-home}\\\textbf{furnishing}}
& \makecell[c]{\textbf{Load-bearing}\\\textbf{walls}}
& \makecell[c]{\textbf{Doors \&}\\\textbf{windows}}
& \makecell[c]{\textbf{Pipes \&}\\\textbf{drainage}} \\
\midrule

\textbf{Floorplan generation}\par
{\scriptsize House-GAN~\cite{House-gan}; Yin et al.~\cite{yin2025floorplan}; HouseMind~\cite{housemind}}
& \xmark
& \cmark
& \xmark
& \xmark
& \xmark
& \xmark \\

\textbf{Scene layout generation}\par
{\scriptsize LayoutGPT~\cite{feng2023layoutgpt}; OptiScene~\cite{yang2025optiscene}; HoGLayout~\cite{hoglayout}; LaviGen~\cite{lavigen}}
& \xmark
& \xmark
& \cmark
& \xmark
& \cmark
& \xmark \\

\textbf{Joint floorplan-layout generation}\par
{\scriptsize HouseDiffusion~\cite{shabani2023housediffusion}; Nestwork~\cite{nestwork}}
& \xmark
& \cmark
& \cmark
& \xmark
& \cmark
& \xmark \\

\textbf{Unified residential design}\par
{\scriptsize RenoLayout (Ours)}
& \cmark
& \cmark
& \cmark
& \cmark
& \cmark
& \cmark \\

\bottomrule
\end{tabularx}
\parbox{\textwidth}{\scriptsize\raggedright
\textbf{Note.} \cmark denotes explicit annotation or supervision, whereas \xmark denotes information not explicitly modeled. Full-home furnishing covers furniture, appliances, storage, and local functional layouts. Building-service annotations cover pipes, drainage points, shafts, and flues.}
\caption{Comparison of data annotations in residential design. Unlike existing settings, RenoLayout provides paired pre-design states with post-design solutions, complemented by full-home furnishing as well as design-critical annotations for structures and building services.}
\label{tab:data_annotation_comparison}
\end{table*}

The gap is consequential because structural decisions interact with scene-level choices. Living--Dining--Kitchen (LDK) integration depends on wall removability, whereas wet-function relocation is constrained by drainage. Storage additions must also preserve circulation and door clearance. Residential design therefore requires concrete operational design points, such as dry--wet separation or room merging, which translate high-level lifestyle descriptions into coordinate-level edits. Without paired design-point supervision, models may reproduce design patterns yet fail to perform valid transformations. Direct LLM generation also risks broken boundaries, collisions, and unintended changes to fixed elements \cite{rudman2025forgotten, feng2023layoutgpt}.

To address this issue, we propose \textbf{RenoLayout}, a unified framework that takes a pre-design residential state and a design-point set as input to generate the corresponding post-design residential solution. A 123-entry \textbf{Design Map} defines the shared vocabulary for structural-renovation and furniture-layout operations. Our difference-grounded, VLM-audited annotation protocol derives case-specific design-point sets from observable before--after edits. The CPT--SFT--RL pipeline progressively learns spatial-code priors, instruction following, and constraint compliance. We further introduce the \textbf{RenoLayout Benchmark}, which contains paired designs with full-home furnishing. Its annotations cover structural elements, openings, pipes, and drainage-related constraints. The benchmark supports a comprehensive evaluation of design-point fulfillment, constraint compliance, living quality, and professional approval.

Our contributions are summarized as follows:

\begin{itemize}
    \item We formulate \textbf{constraint-aware unified residential design} as the transformation of an existing home into a coordinated structural and interior solution. To support this task, we introduce the \textbf{RenoLayout Benchmark}, which contains 200 paired residential-design cases with full-home furnishing and building-service annotations. It will be publicly released.

    \item We propose a 123-entry \textbf{Design Map} together with a difference-grounded, VLM-audited protocol that converts observable transformations into controllable supervision.

    \item We present \textbf{RenoLayout}, an executable \textit{Geometry-as-Code} framework trained through a staged CPT--SFT--RL pipeline. Our evaluation covers four complementary dimensions: design-point fulfillment, constraint compliance, living quality, and professional approval.
\end{itemize}

\section{Related Work}

\subsection{Floorplan Generation}

Floorplan generation has progressed from raster-based GANs \cite{House-gan,nauata2021house} to diffusion and discrete-denoising methods that improve geometric controllability and topological consistency \cite{shabani2023housediffusion,chen2023polydiffuse,qin2024chathousediffusionpromptguidedgenerationediting,su2024floor,zeng2024residential}. Recent LLM-based approaches enable language-guided refinement, next-room prediction, bubble-diagram augmentation, and multimodal tokenization \cite{luo2024dstruct2design,li2024chatdesign,yin2025floorplan,wei2024graph,housemind}; concurrent work uses reinforcement learning to enforce numerical and connectivity constraints, while GRE-Diff supports constraint-aware vector generation and iterative editing \cite{lara2026floorplanrlvr,wang2026grediff}.

\subsection{Scene Generation }

LLMs have been adapted as visual planners that combine semantic planning with geometric reasoning. LayoutGPT \cite{feng2023layoutgpt} targets 2D object placement, while Holodeck and OptiScene extend language-guided planning to 3D scene construction \cite{yang2024holodeck, yang2025optiscene}. Recent methods improve 3D layouts through hierarchical VLM-guided optimization and autoregressive generation \cite{hoglayout, lavigen}. Architect-Ant instead targets editable furnishing of architectural floorplans through a coordinate-based DSL and preference optimization \cite{rodionov2026architectant}. More directly, Co-Layout jointly optimizes room and furniture layouts from LLM-derived constraints, while HomeWorld connects whole-home floorplan generation with hierarchical furnishing \cite{xiang2025colayout, li2026homeworld}.

Although Co-Layout and HomeWorld move toward the joint generation of spatial structures and furniture layouts, they synthesize designs from requirements or generated floorplans instead of learning transformations from paired before--after designs \cite{xiang2025colayout, li2026homeworld}. Architect-Ant assumes a fixed architectural shell and addresses furnishing without structural renovation \cite{rodionov2026architectant}. Their data primarily capture final solutions and therefore provide no explicit before--after supervision for operational renovation actions such as merging rooms, relocating wet functions, or preserving fixed services. In contrast, our setting begins with a real pre-design residential state and requires coordinated modification of partitions, room functions, and furniture while preserving load-bearing walls, openings, shafts, and service locations.
\subsection{Spatial Reasoning}
To align semantic generation with geometric precision, recent approaches incorporate explicit constraints ranging from differentiable optimization \cite{sun2025layoutvlm} to parametric generation \cite{wang20252d}. These mechanisms address ambiguities in model reasoning, where perspectives diverge: while \cite{ivanitskiy2023structured} suggest models possess emergent topological representations, \cite{rudman2025forgotten} argue they lack innate visual reasoning, necessitating structured guidance.
\section{RenoLayout}

\subsection{Task and Data Construction}
\label{sec:task_data}

To align our task formulation with real-world residential design practice, where design begins with an existing home rather than an empty boundary, we formulate residential design as a \emph{design-point-guided transformation} task instead of a from-scratch generation task. Given a complete pre-design residential state and a design-point set, our goal is to generate the corresponding post-design residential solution, comprising a renovated floorplan and full-home furniture layout, while preserving immutable structural elements and modifying editable components according to the specified design objectives.

Let \(X^{pre}\) denote the pre-design residential state and \(X^{post}\) denote the corresponding post-design residential solution. Both are represented as sets of geometric and semantic entities, including walls, doors, windows, rooms and furniture. We decompose the pre-design state into immutable structural elements and editable elements:
\begin{equation}
    X^{pre} = S \cup E^{pre}, \qquad S \cap E^{pre}=\emptyset,
\end{equation}
where \(S\) denotes immutable structural elements, such as exterior boundaries, load-bearing walls and shafts, and \(E^{pre}\) denotes editable elements, such as non-load-bearing partitions, room functions, movable door openings, furniture and local scene arrangements.

\noindent Let the design-point set be
\begin{equation}
    \mathcal{D} = \{d_1, d_2, \ldots, d_k\},
\end{equation}
The desired output is an executable design solution \(\hat{X}^{post}\) that balances model likelihood and design-point fulfillment under hard preservation and validity constraints:
\begin{equation}
\label{eq:task_objective}
\begin{aligned}
\hat{X}^{post} = \arg\max_{X \in \mathcal{X}} \;&
\log P_{\theta}(X \mid X^{pre}, \mathcal{D})
+ \lambda G(\mathcal{D},X) \\
\text{s.t.}\;& \Pi_S(X)=S, \qquad V(X)=1,
\end{aligned}
\end{equation}
where \(\mathcal{X}\) denotes the space of possible post-design solutions. The operator \(\Pi_S\) extracts the immutable structural and building-service elements from a candidate design \(X\). The constraint \(\Pi_S(X)=S\) enforces exact preservation of their geometry and attributes. The binary predicate \(V(X)\) indicates whether \(X\) satisfies all other hard feasibility requirements, such as valid geometry, room connectivity, and collision-free furniture placement. We define normalized design-point fulfillment as
\begin{equation}
    G(\mathcal{D},X)=\frac{1}{|\mathcal{D}|}
    \sum_{d\in\mathcal{D}}\mathbb{I}[d\text{ is fulfilled in }X],
\end{equation}
where \(G(\mathcal{D},X)\in[0,1]\), and \(\lambda\geq0\) controls its contribution relative to model likelihood. Equation~\ref{eq:task_objective} is a task-level specification rather than an inference-time optimization algorithm. 

This formulation differs from prior floorplan generation and scene synthesis tasks in two important aspects. First, the input is a complete pre-design residential state rather than an empty boundary, a partial skeleton, or a text-only prompt. This makes the task closer to practical residential design workflows, where designers transform an existing home into an improved solution. Second, the control signal comprises professional design points distilled from real-world design cases and consolidated with the expertise of human interior designers, rather than free-form user requirements or personas. This enables explicit supervision, category-wise evaluation, and difference-grounded annotation from paired before--after design data.

\subsection{Design Map}
\label{sec:design_point_map}
\begin{figure}[t]
    \centering
    \includegraphics[width=0.9\linewidth]{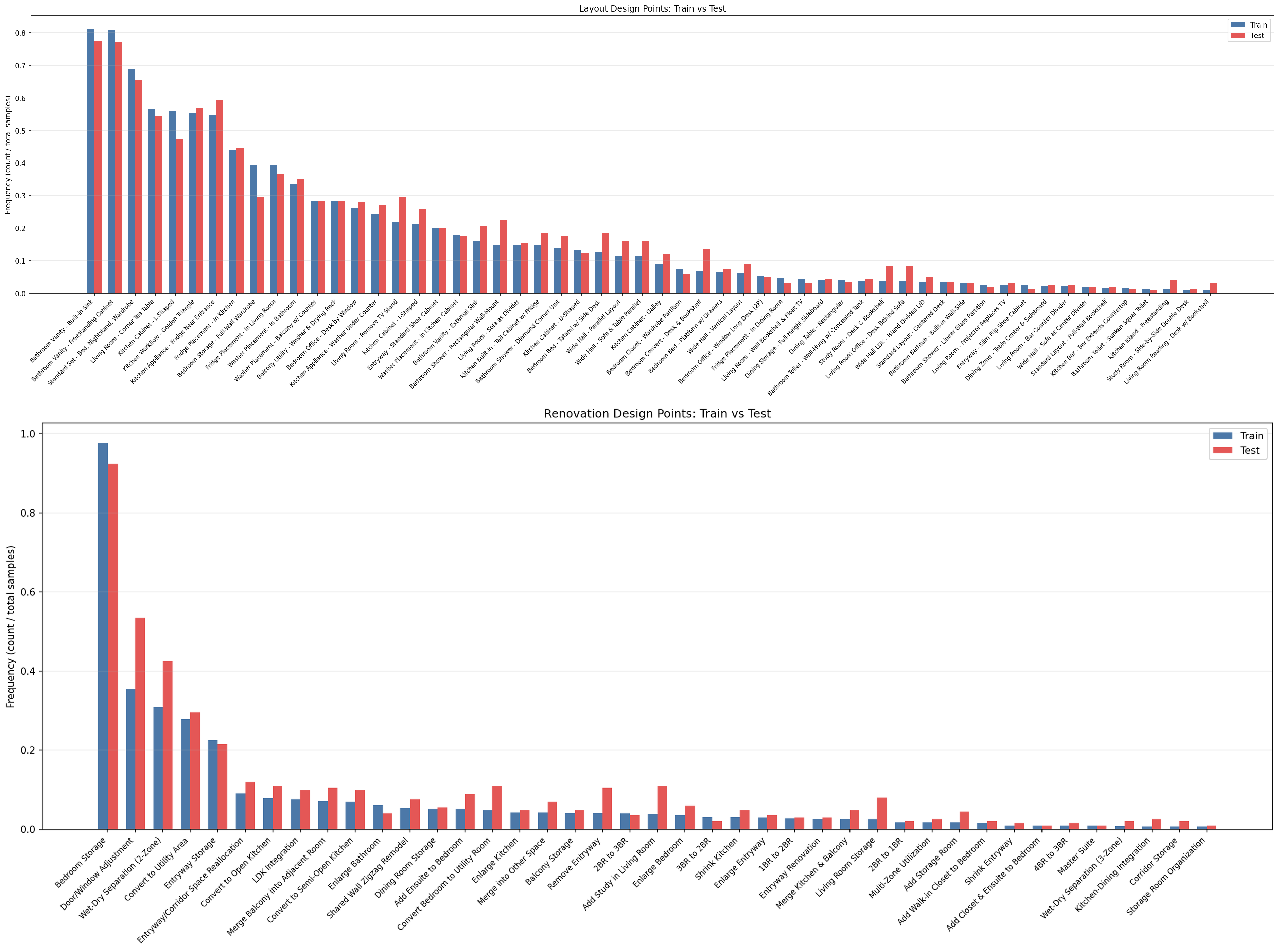}
    \caption{Frequency of furniture-layout (top) and structural-renovation (bottom) design points in the training (blue) and test (red) splits. Frequency is the occurrence count normalized by the total number of samples per split; design points are ordered by descending training-set frequency.}
    \label{fig:RenoLayout_stats}
\end{figure}
\begin{figure}[t]
    \centering
    \includegraphics[width=0.95\linewidth]{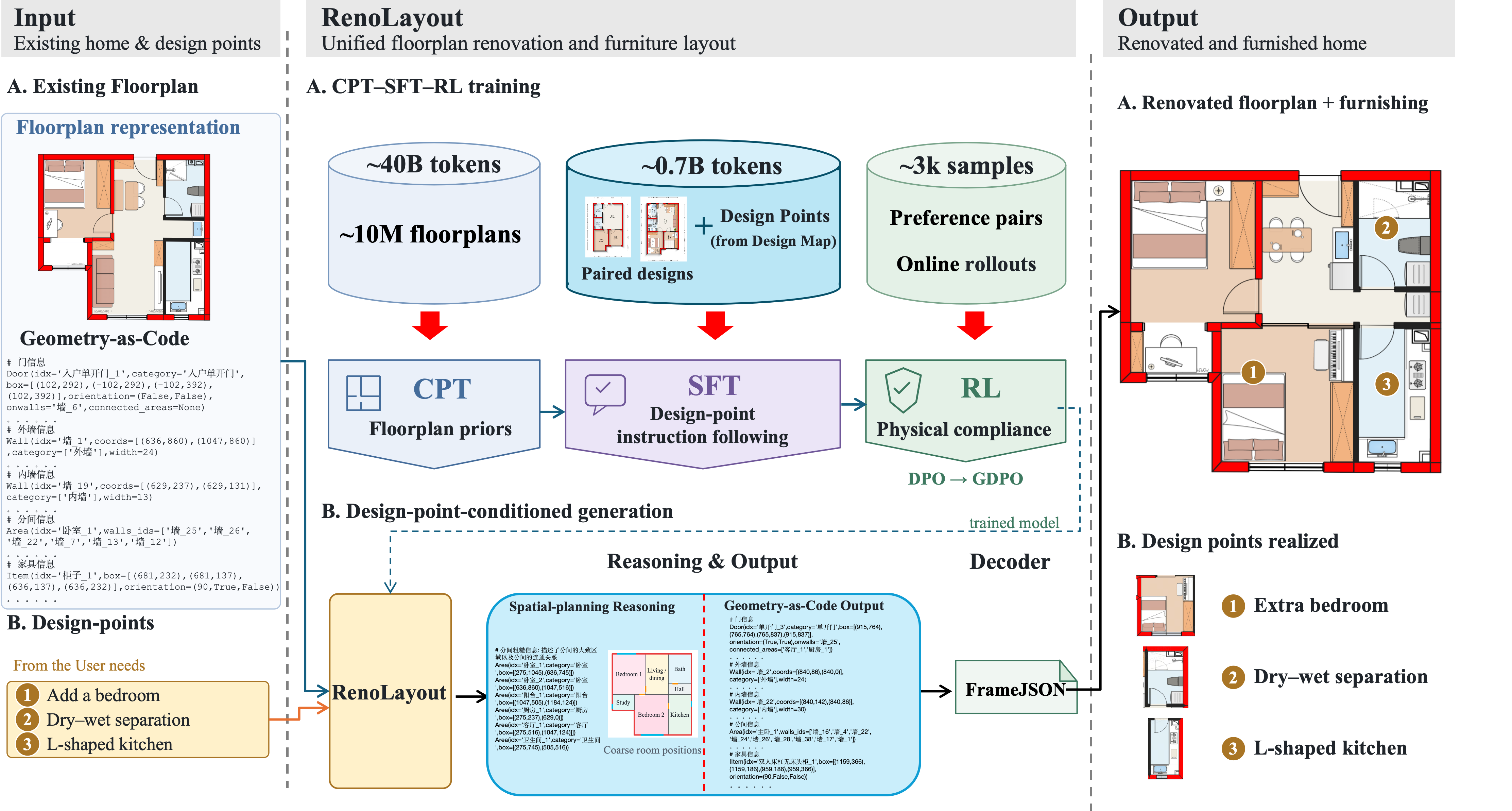}
    \caption{Overview of RenoLayout. Left: paired designs are converted into difference-grounded design-point sets and verified through multi-VLM consistency auditing. Middle: the pre-design state is serialized as Geometry-as-Code and combined with design-point prompts, enabling Qwen3 to first produce a graph-based spatial plan and then generate executable structural-renovation and furniture-layout actions through the CPT--SFT--RL training pipeline.}
    \label{fig:method}
\end{figure}
Notably, the user requirements are often high-level, ambiguous, and expressed in non-technical language, making them unsuitable as direct operational controls for residential design. In professional practice, interior designers translate such requirements into concrete and actionable design decisions. Inspired by this professional practice, we abstract these decisions as design points and organize them into a hierarchical \emph{Design Map}, which serves as the model's control vocabulary instead of raw user requirements. Each fine-grained design point corresponds to a specific structural-renovation or furniture-layout operation.


We construct the \emph{Design Map} by consolidating recurring transformations observed in paired design data with the domain expertise of professional interior designers. The resulting taxonomy comprises 123 fine-grained design points: 52 structural-renovation points and 71 furniture-layout points. Together, they cover both floorplan-level transformations and scene-level arrangements. Representative examples include LDK integration, dry--wet bathroom separation, bedroom-to-closet conversion, entrance storage, appliance relocation, and furniture reachability optimization.


The Design Map is not used as a rule-based generator. Instead, it defines the control vocabulary, annotation schema, and evaluation interface for the transformation task. RenoLayout learns from paired before--after examples to synthesize complete post-design Geometry-as-Code programs conditioned on selected design points.

\subsection{Difference-Grounded Design-Point Annotation}
\label{sec:difference_grounded_annotation}

Given paired before--after design data, we construct design-point labels from observable transformations rather than from free-form user descriptions. For each paired example \((X^{pre}, X^{post})\), we first compute a structured difference:
\begin{equation}
    \Delta X = \mathrm{Diff}(X^{pre}, X^{post}).
\end{equation}
The difference \(\Delta X\) captures geometric and semantic edits at multiple levels, such as wall addition or removal, room merging or splitting, room-function reassignment, and furniture relocation.

We then map the extracted differences to entries in the Design Map through deterministic rules:
\begin{equation}
    \mathcal{D} = R(\Delta X, X^{pre}, X^{post}),
\end{equation}
where \(R\) denotes the rule-grounded mapper. Each rule links an observable before--after edit pattern to one or more fine-grained design points. For example, if the kitchen partition is removed and the living, dining, and kitchen areas become connected, the pair is mapped to the LDK integration design point.

This rule-grounded mapping produces the candidate design-point set \(\mathcal{D}\) directly from observable before--after edits rather than free-form descriptions.


\subsection{Annotation Audit and Quality Control}
\label{sec:annotation_quality}

We then use multiple VLMs solely as annotation auditors. Given the pre-design image, post-design image, extracted differences, and rule-derived labels, the auditors identify labels that are inconsistent with the observed edits, mismatched with the transformation type, duplicated, or conflicting.

Labels validated by the auditors are retained, while unsupported or ambiguous cases are removed or manually reviewed.

\subsection{Triplet Data Construction and RenoLayout Benchmark}
\label{sec:paired_dataset}

Using the audited design-point labels, we construct supervised transformation triplets:
\begin{equation}
    (X^{pre}, \mathcal{D}, X^{post}),
\end{equation}
Each sample contains floorplan-level information, including room boundaries, room functions, doors, windows, and structural elements, as well as scene-level information, including furniture, appliances, cabinets, and functional objects. Both \(X^{pre}\) and \(X^{post}\) are serialized into the same Geometry-as-Code representation, allowing the model to learn transformations over a unified executable spatial program. The RenoLayout Benchmark supports reproducible evaluation and contains 200 paired real-world residential-design cases that are strictly disjoint from all data used for CPT, SFT, and RL. The post-design solutions were produced by professional interior designers and are used as ground-truth designs. To ensure reliable and comprehensive design-point coverage, we manually reviewed and filtered all cases. Figure~\ref{fig:RenoLayout_stats} visualizes the frequency distribution of individual design points, while Table~\ref{tab:test_design_point_distribution} provides complementary benchmark-level statistics, including the total numbers of structural-renovation and furniture-layout annotations and their per-case distributions.

\begin{table}[t]
    \centering
    \scriptsize
    \setlength{\tabcolsep}{2.5pt}
    \begin{tabular*}{0.95\columnwidth}{@{\extracolsep{\fill}}lccccc@{}}
        \toprule
        \textbf{Category} & \textbf{Total} & \textbf{Avg./case} & \textbf{Std.} & \textbf{Min} & \textbf{Max} \\
        \midrule
        Renovation & 875 & 4.38 & 2.33 & 1 & 13 \\
        Layout & 2,177 & 10.88 & 3.40 & 0 & 23 \\
        \midrule
        \textbf{Overall} & \textbf{3,052} & \textbf{15.26} & \textbf{4.71} & \textbf{2} & \textbf{31} \\
        \bottomrule
    \end{tabular*}
    \caption{Per-case statistics and total counts of structural-renovation and furniture-layout design-point annotations across the 200-case RenoLayout Benchmark.}
    \label{tab:test_design_point_distribution}
\end{table}


\subsection{Continual Pre-Training (CPT)}

General-purpose LLMs often struggle to comprehend the specialized vector representations required for residential design. To bridge the domain gap, we employ CPT, which effectively aligns the model with these geometric code constraints~\cite{qi2025evolm}.

We conduct CPT on approximately 40 billion tokens, centered on vector-code data derived from roughly 10 million floorplans. The corpus incorporates two key strategies: Geometry-as-Code, which represents floorplans as Python code to leverage the spatial reasoning of LLMs, and a \textbf{Data Mixing Strategy} that combines mathematical~\cite{fujii2025rewritingpretrainingdataboosts} and general-purpose data to mitigate catastrophic forgetting. See the supplementary material for details.

\subsection{Supervised Fine-Tuning (SFT) with Feature Enhancement}
Although CPT equips the model with geometric priors, it does not fully align generation with complex design semantics. We therefore perform SFT on approximately 34,000 real-world floorplans, expanded to approximately 130,000 samples (0.7 billion tokens) through rotational augmentation. Additional data details are provided in the supplementary material.
\label{Structured_Feature_Enhancement}

\noindent\textbf{Bounding-Box Encoding.}
To improve local geometric control, we replace the center-size representations of furniture and doors with explicit four-corner bounding boxes. Exposing object boundaries directly makes inter-object clearance, collision avoidance, and doorway occlusion easier to learn.

\noindent\textbf{Graph-Planning CoT.}
To reduce topological and sizing errors, we adopt a coarse-to-fine planning process. The model first predicts a topological adjacency graph (Bubble Diagram~\cite{House-gan,luo2024dstruct2design}) to establish room-level spatial logic, and then generates detailed spatial actions conditioned on this structural plan.

\subsection{Reinforcement Learning}
\label{sec:method_rl}

\noindent\textbf{Off-policy Preference Optimization via DPO.}
We first apply DPO~\cite{rafailov2023direct} to establish an offline preference guardrail covering both physical validity and instruction following. We construct fixed chosen--rejected pairs through a two-stage selection procedure. In the first stage, the ground-truth design $y_{\mathrm{GT}}$ serves as the chosen upper-bound anchor, while a predicted candidate $y$ is eligible as rejected only if it fails physical quality control and achieves lower design-point recall than $y_{\mathrm{GT}}$. Among eligible candidates, we select the one with the smallest length difference from the ground truth, reducing spurious preferences based on output length.

In the second stage, we form fine-grained preference pairs among predicted candidates. For an output $y$, let $e(y)$ denote the number of detected violations. We define its physical-validity reward uniformly as $R_{phys}(y)=\max(0,1-e(y)/10)$, which equals 1 when all quality-control checks pass. We rank candidates using
\begin{equation}
    S(y)=\omega_{phys}R_{phys}(y)+\omega_{inst}R_{inst}(y),
\end{equation}
where $R_{inst}(y)$ is design-point recall. We set the physical-validity and instruction-following weights to $\omega_{phys}=0.8$ and $\omega_{inst}=0.2$, respectively. A pair is retained only when its score advantage is at least $0.3$ and its length difference is no more than $15\%$ of the mean pair length. We then use Hungarian matching to maximize the total preference advantage across non-redundant pairs. This construction emphasizes the weaker physical-validity dimension while preserving instruction following and filtering noisy or length-biased comparisons.

\noindent\textbf{On-policy Optimization via GDPO.}
Residential design is inherently a multi-objective optimization problem involving trade-offs among format correctness, instruction following, physical validity, and item proportion. These objectives differ substantially in reward scale, variance, and optimization difficulty. Standard GRPO~\cite{shao2024deepseekmath} aggregates multiple reward components before group-wise normalization, which can collapse distinct reward combinations into similar advantages and allow high-variance components to dominate the policy update, suppressing informative signals from harder objectives. We therefore adopt GDPO~\cite{liu2026gdpo}, which independently normalizes each reward component before aggregating their advantages. This reward-decoupled formulation preserves objective-specific learning signals and enables more balanced optimization across competing residential-design objectives.

Using samples generated from the current policy, we apply direct rule-based feedback to format correctness, instruction following, physical validity, and item proportion. For each sampled output $i$, GDPO independently converts each reward component into a group-normalized advantage:
\begin{equation}
    A_{i,k} = \frac{R_{i,k}-\mu_k^{\mathcal{G}}}
    {\sigma_k^{\mathcal{G}}+\epsilon},
    \quad k\in\mathcal{K},
\end{equation}
where $\mathcal{K}=\{\mathrm{form},\mathrm{inst},\mathrm{phys},\mathrm{prop}\}$, and $\mu_k^{\mathcal{G}}$ and $\sigma_k^{\mathcal{G}}$ are the mean and standard deviation of reward component $k$ within rollout group $\mathcal{G}$. The final advantage used for policy optimization is
\begin{equation}
    A_i^{\mathrm{GDPO}}
    = \sum_{k\in\mathcal{K}} \omega_k A_{i,k}.
\end{equation}

The four reward components respectively evaluate output-format correctness ($R_{form}$), design-point recall ($R_{inst}$), physical validity ($R_{phys}$), and item proportion ($R_{prop}$). We set $\omega_{form}=\omega_{phys}=\omega_{prop}=1.0$ and $\omega_{inst}=0.5$. Because the physical objectives are more difficult to optimize, the lower instruction-following weight prevents the policy from overemphasizing this objective at the expense of physical validity and appropriate item configuration.

\section{Experiments}
\subsection{Metrics and Training Details}

\noindent\textbf{Evaluation setting. }
We evaluate all methods on the RenoLayout Benchmark. For each case $i$, a method receives a pre-design residential state $X_i^{pre}$ and a design-point set $\mathcal{D}_i$, and produces a complete post-design residential solution $\hat{X}_i^{post}$. The evaluation addresses four complementary questions: whether the requested design points are fulfilled, whether the transformation satisfies design and physical constraints, whether the resulting home provides high living quality, and whether the overall design receives professional approval.

\noindent\textbf{Design-Point Fulfillment. }
We evaluate instruction following through design-point fulfillment using Recall@1. Let $z_{i,k}^{(\mathrm{rule})}\in\{0,1\}$ indicate whether the rule-based evaluator judges the $k$-th requested design point in case $i$ to be realized in the first generated post-design solution, and let $M_i=|\mathcal{D}_i|$ denote the number of requested design points for that case. Recall@1 first computes the proportion of fulfilled design points within each case and then averages these case-level recalls over all $N=200$ cases:
\begin{equation}
\mathrm{Recall@1}=\frac{100}{N}\sum_{i=1}^{N}
\frac{1}{M_i}\sum_{k=1}^{M_i}z_{i,k}^{(\mathrm{rule})}.
\end{equation}

\noindent\textbf{Constraint Compliance (Pass@1). }
We use 22 binary checks grouped into \textit{Geometric Validity}, \textit{Structural Compliance}, \textit{Functional Integrity}, and \textit{Collision-Free Placement}. Let $q_{i,m}\in\{0,1\}$ indicate whether case $i$ passes check $m$. Each category rate is the percentage of cases that pass every check in that category, while Pass@1 requires all 22 checks to pass simultaneously:
\begin{equation}
\mathrm{Pass@1}=\frac{100}{N}\sum_{i=1}^{N}\mathbb{I}\!\left[\sum_{m=1}^{22}q_{i,m}=22\right].
\end{equation}
Failure of any single check makes the entire case fail. All five quantities are percentages, and higher values are better.

\noindent\textbf{VLM-Based Habitability Score (VLM-HS). }
We use four independent VLM evaluators: Seed2.1 Pro~\cite{doubao_seed21}, GLM-5V-Turbo~\cite{glm5team2026glm5vibecodingagentic}, GPT-5.5~\cite{gpt55}, and Gemini-3.1-Flash-Lite~\cite{gemini31flashlite}. Each separately scores functional zoning, spatial layout, circulation quality, amenity configuration, design-point implementation, and physical compliance. For each evaluator, the first five dimensions form a weighted subjective score, while physical compliance is fused with a deterministic checker for P0/P1 violations. Predictions and ground-truth plans are scored independently before case-matched deficit calibration, yielding four model-specific VLM-HS values without cross-model averaging. Complete formulas, parameters, rubrics, and prompts are provided in the supplementary material.

\noindent\textbf{Professional evaluation protocol.}
To assess professional approval alongside the automatic metrics, three interior designers independently evaluated anonymized and randomly ordered RenoLayout outputs and corresponding ground-truth designs for all 200 cases using the Professional Approval Score (PAS; 0--10); scoring details are provided in the supplementary material.

\noindent\textbf{Training Details. }
Our model is built upon Qwen3-8B~\cite{yang2025qwen3}.
In the CPT stage, the model was trained for 2 epochs with a global batch size of 32 and a learning rate of $1\times10^{-5}$. Subsequently, during SFT, we reduced the batch size to 16 and the learning rate to $5\times10^{-6}$, training for 2 epochs. For the reinforcement learning stages, we train GDPO/DPO for 2 epochs with a global batch size of 64 and a learning rate of $3\times10^{-6}$. To maintain training stability, we set the group size to 8, the KL coefficient to 0.05, and the sampling temperature to 0.6.

\subsection{Feature Analysis}
\label{sec:ablation_feat}

\begin{table}[t]
    \centering
    \scriptsize
    \setlength{\tabcolsep}{1.5pt}
    \renewcommand{\arraystretch}{0.96}
    \begin{tabular*}{0.95\linewidth}{@{\extracolsep{\fill}}ccc cccccc@{}}
        \toprule
        \multicolumn{3}{c}{\textbf{Settings}} & \multicolumn{2}{c}{\textbf{Overall Metrics}} & \multicolumn{4}{c}{\textbf{VLM-HS by Evaluator}} \\
        \cmidrule(r){1-3} \cmidrule(lr){4-5} \cmidrule(l){6-9}
        \textbf{CPT} & \textbf{Box} & \textbf{CoT} &
        \textbf{Recall@1}$\uparrow$ & \textbf{Pass@1}$\uparrow$ & \textbf{Seed} & \textbf{GLM} & \textbf{GPT} & \textbf{Gemini} \\
        \midrule
        \xmark & \xmark & \xmark & 80.92\% & 27.00\% & 5.12 & 4.97 & 5.60 & 5.07 \\
        \xmark & \cmark & \xmark & 81.13\% & 38.50\% & 5.86 & 5.85 & 6.39 & 5.77 \\
        \xmark & \cmark & \cmark & 84.47\% & 42.00\% & 5.98 & 5.86 & 6.46 & 5.78 \\
        \cmark & \xmark & \cmark & 88.65\% & 31.50\% & 6.32 & 6.01 & 6.69 & 6.19 \\
        \cmark & \cmark & \xmark & 86.37\% & 45.00\% & 6.61 & 6.60 & 7.10 & 6.40 \\
        \cmark & \cmark & \cmark & \textbf{88.67}\% & \textbf{49.00}\% & \textbf{6.82} & \textbf{6.75} & \textbf{7.22} & \textbf{6.45} \\
        \midrule
        \multicolumn{3}{c}{\textbf{Settings}} & \multicolumn{6}{c}{\textbf{Physical Pass Rates}} \\
        \cmidrule(r){1-3} \cmidrule(l){4-9}
        \textbf{CPT} & \textbf{Box} & \textbf{CoT} &
        $\mathcal{P}_{\mathrm{geom}}\uparrow$ & \multicolumn{2}{c}{$\mathcal{P}_{\mathrm{struct}}\uparrow$} & \multicolumn{2}{c}{$\mathcal{P}_{\mathrm{func}}\uparrow$} & $\mathcal{P}_{\mathrm{coll}}\uparrow$ \\
        \midrule
        \xmark & \xmark & \xmark & 93.00\% & \multicolumn{2}{c}{79.00\%} & \multicolumn{2}{c}{89.00\%} & 32.00\% \\
        \xmark & \cmark & \xmark & 92.00\% & \multicolumn{2}{c}{80.50\%} & \multicolumn{2}{c}{89.50\%} & 49.00\% \\
        \xmark & \cmark & \cmark & 95.00\% & \multicolumn{2}{c}{78.00\%} & \multicolumn{2}{c}{94.00\%} & 53.00\% \\
        \cmark & \xmark & \cmark & 97.50\% & \multicolumn{2}{c}{79.00\%} & \multicolumn{2}{c}{95.50\%} & 39.00\% \\
        \cmark & \cmark & \xmark & 97.50\% & \multicolumn{2}{c}{76.50\%} & \multicolumn{2}{c}{95.50\%} & \textbf{57.00}\% \\
        \cmark & \cmark & \cmark & \textbf{98.00}\% & \multicolumn{2}{c}{\textbf{81.00}\%} & \multicolumn{2}{c}{\textbf{96.50}\%} & \textbf{57.00}\% \\
        \bottomrule
    \end{tabular*}
    \caption{Feature ablation of CPT, bounding-box encoding (Box), and graph-planning CoT. All metrics except VLM-HS (0--10) are percentages.}
    \label{tab:ablation_study}
\end{table}
Table~\ref{tab:ablation_study} incrementally introduces bounding-box features and CoT with and without CPT. The comparison tests whether domain pre-training improves design-point fulfillment, constraint compliance, and living quality, whether explicit bounding boxes strengthen local geometric control, and whether CoT benefits global planning and final living quality.

\begin{figure}[!t]
    \centering
    \includegraphics[width=0.9\linewidth]{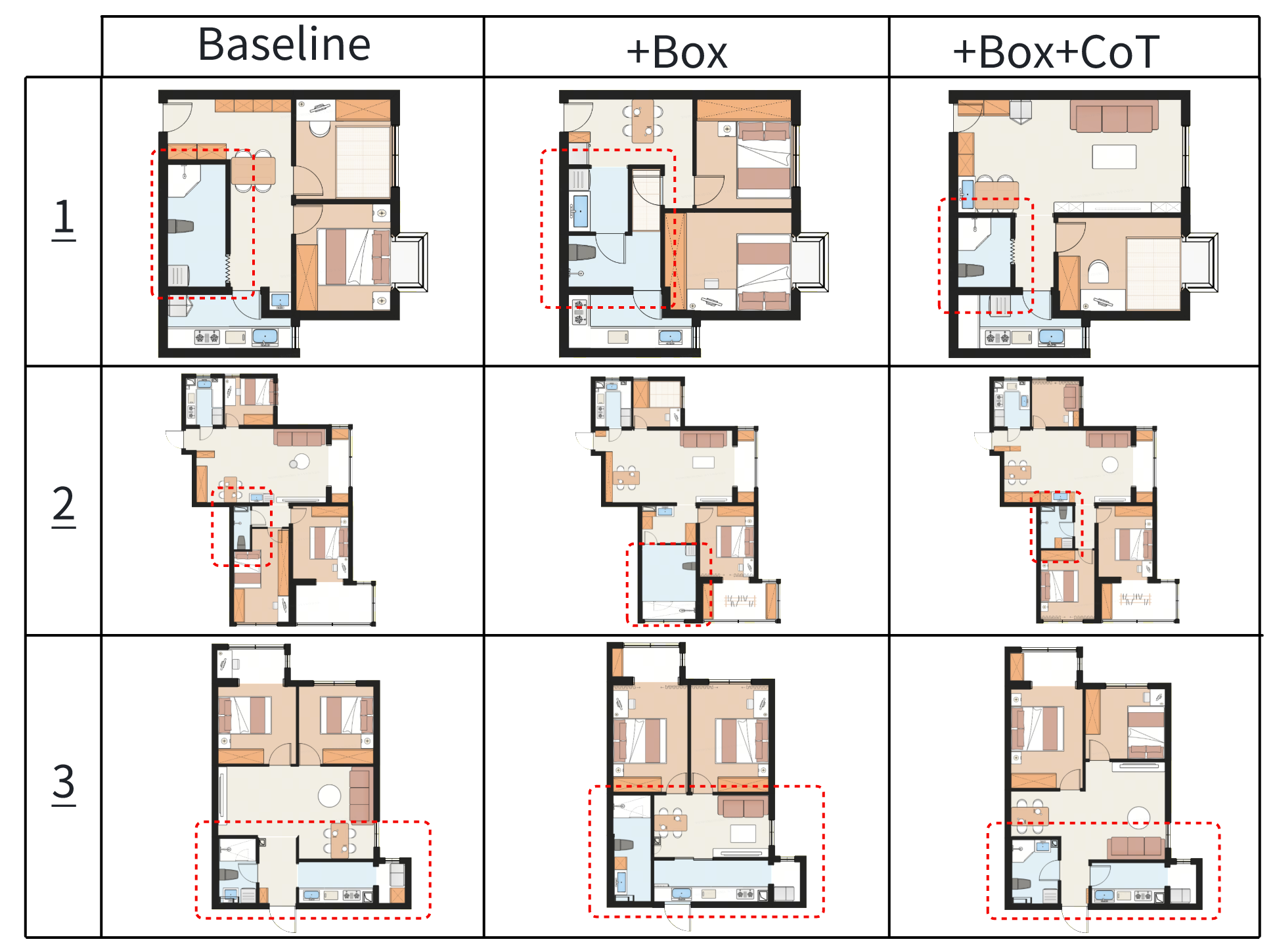}
    \caption{Qualitative effects of bounding boxes on local placement and CoT on spatial planning.}
    \label{fig:feature}
\end{figure}

The components provide complementary benefits with distinct tradeoffs. Without CPT, adding CoT to the bounding-box representation increases Recall@1 from 81.13\% to 84.47\%, Pass@1 from 38.50\% to 42.00\%, and collision-free placement from 49.00\% to 53.00\%, although structural compliance decreases from 80.50\% to 78.00\%. Under CPT, adding CoT raises Recall@1 from 86.37\% to 88.67\%, Pass@1 from 45.00\% to 49.00\%, and structural compliance from 76.50\% to 81.00\%, while collision-free placement remains at 57.00\%. The four VLM-HS scores, ordered as Seed, GLM, GPT, and Gemini, increase from $(6.61, 6.60, 7.10, 6.40)$ to $(6.82, 6.75, 7.22, 6.45)$. Compared with the corresponding Box+CoT model without CPT, the complete configuration improves Recall@1 and Pass@1 by 4.20 and 7.00 percentage points, respectively. These results suggest that CoT primarily strengthens global design-point fulfillment and overall constraint compliance, while bounding boxes support local placement quality and CPT supplies complementary domain priors. Figure~\ref{fig:feature} further provides representative visual evidence for these complementary effects.

\subsection{Ablation on Different RL Strategies: DPO vs. GDPO}
\label{sec:rl_ablation}
\begin{table}[t]
    \centering
    \scriptsize
    \setlength{\tabcolsep}{1.5pt}
    \renewcommand{\arraystretch}{0.96}
    \begin{tabular*}{0.95\columnwidth}{@{\extracolsep{\fill}}lcccccc@{}}
        \toprule
        \textbf{Method} & \multicolumn{2}{c}{\textbf{Overall Metrics}} & \multicolumn{4}{c}{\textbf{VLM-HS by Evaluator}} \\
        \cmidrule(r){2-3} \cmidrule(l){4-7}
        & \textbf{Recall@1}$\uparrow$ & \textbf{Pass@1}$\uparrow$ & \textbf{Seed} & \textbf{GLM} & \textbf{GPT} & \textbf{Gemini} \\
        \midrule
        RenoLayout-SFT & 88.67\% & 49.00\% & 6.82 & 6.75 & 7.22 & 6.45 \\
        + DPO & 89.37\% & 55.50\% & 6.92 & 6.90 & 7.36 & 6.64 \\
        + GDPO & 90.89\% & 53.00\% & 6.87 & 6.81 & 7.20 & 6.77 \\
        + DPO + GDPO & \textbf{91.63}\% & 61.00\% & \textbf{7.26} & \textbf{7.32} & \textbf{7.72} & \textbf{7.22} \\
        + GDPO + DPO & 90.42\% & \textbf{61.50}\% & 7.17 & 7.23 & 7.70 & 7.16 \\
        \midrule
        \textbf{Method} & \multicolumn{6}{c}{\textbf{Physical Pass Rates}} \\
        \cmidrule(l){2-7}
        & $\mathcal{P}_{\mathrm{geom}}\uparrow$ & \multicolumn{2}{c}{$\mathcal{P}_{\mathrm{struct}}\uparrow$} & \multicolumn{2}{c}{$\mathcal{P}_{\mathrm{func}}\uparrow$} & $\mathcal{P}_{\mathrm{coll}}\uparrow$ \\
        \midrule
        RenoLayout-SFT & 98.00\% & \multicolumn{2}{c}{81.00\%} & \multicolumn{2}{c}{\textbf{96.50}\%} & 57.00\% \\
        + DPO & 98.00\% & \multicolumn{2}{c}{84.50\%} & \multicolumn{2}{c}{96.00\%} & 63.50\% \\
        + GDPO & \textbf{99.00}\% & \multicolumn{2}{c}{80.50\%} & \multicolumn{2}{c}{95.50\%} & 66.50\% \\
        + DPO + GDPO & \textbf{99.00}\% & \multicolumn{2}{c}{85.00\%} & \multicolumn{2}{c}{\textbf{96.50}\%} & 69.50\% \\
        + GDPO + DPO & 98.50\% & \multicolumn{2}{c}{\textbf{85.50}\%} & \multicolumn{2}{c}{96.00\%} & \textbf{70.00}\% \\
        \bottomrule
    \end{tabular*}
    \caption{RL post-training ablation of DPO and GDPO. All metrics except VLM-HS (0--10) are percentages.}
    \label{tab:rl_ablation}
\end{table}

Table~\ref{tab:rl_ablation} compares offline (DPO) and online (GDPO) optimization and highlights their different effects on residential design. DPO yields a larger improvement in structural compliance, indicating that preference supervision from fixed comparison pairs primarily benefits global structural consistency. In contrast, GDPO produces a larger improvement in collision-free placement, consistent with its online optimization objective that directly incorporates geometric feedback during policy refinement, while leading to a slight reduction in functional integrity. These observations suggest that DPO mainly improves global structural validity, whereas GDPO is more effective at correcting instruction-specific geometric errors.

The results further indicate that the optimization order influences the final policy. Applying DPO before GDPO achieves the highest Recall@1 (91.63\%), together with 61.00\% Pass@1 and the highest VLM-HS across Seed, GLM, GPT, and Gemini (7.26, 7.32, 7.72, and 7.22, respectively). Reversing the order yields the highest Pass@1 (61.50\%), structural compliance (85.50\%), and collision-free placement (70.00\%), but reduces Recall@1 to 90.42\% and yields lower scores on all four VLM-HS evaluators. Based on its stronger design-point fulfillment and overall balance, we adopt the DPO--GDPO variant as the final \textbf{RenoLayout} model and denote its pre-RL checkpoint as \textbf{RenoLayout-SFT}.

\subsection{Professional Evaluation}

In the blinded evaluation (Table~\ref{tab:pas}), three professional interior designers rated RenoLayout outputs at 5.43 on average, compared with 7.00 for the corresponding designer-produced ground-truth layouts.

\begin{table}[t]
    \centering
    \scriptsize
    \setlength{\tabcolsep}{3.2pt}
    \begin{tabular*}{0.95\columnwidth}{@{\extracolsep{\fill}}lccccc@{}}
        \toprule
        \textbf{Method} & \textbf{Designer 1} & \textbf{Designer 2} & \textbf{Designer 3} & \textbf{Mean}$\uparrow$ & \textbf{Std.}$\downarrow$ \\
        \midrule
        RenoLayout & 5.50 & 5.59 & 5.19 & 5.43 & 0.17 \\
        Ground Truth (GT) & 7.06 & 7.35 & 6.58 & 7.00 & 0.32 \\
        \bottomrule
    \end{tabular*}
    \caption{Professional Approval Score (PAS; 0--10).}
    \label{tab:pas}
\end{table}

\subsection{Qualitative Comparison with LayoutGPT}


We also compare room-level furniture arrangements with LayoutGPT~\cite{feng2023layoutgpt}. Following its standard setting, the comparison uses one bedroom and one living room per sample. The examples complement the unified evaluation by illustrating furniture reachability, wall alignment, and collision-related errors.
\begin{figure}[!t]
    \centering
    \includegraphics[width=0.9\linewidth]{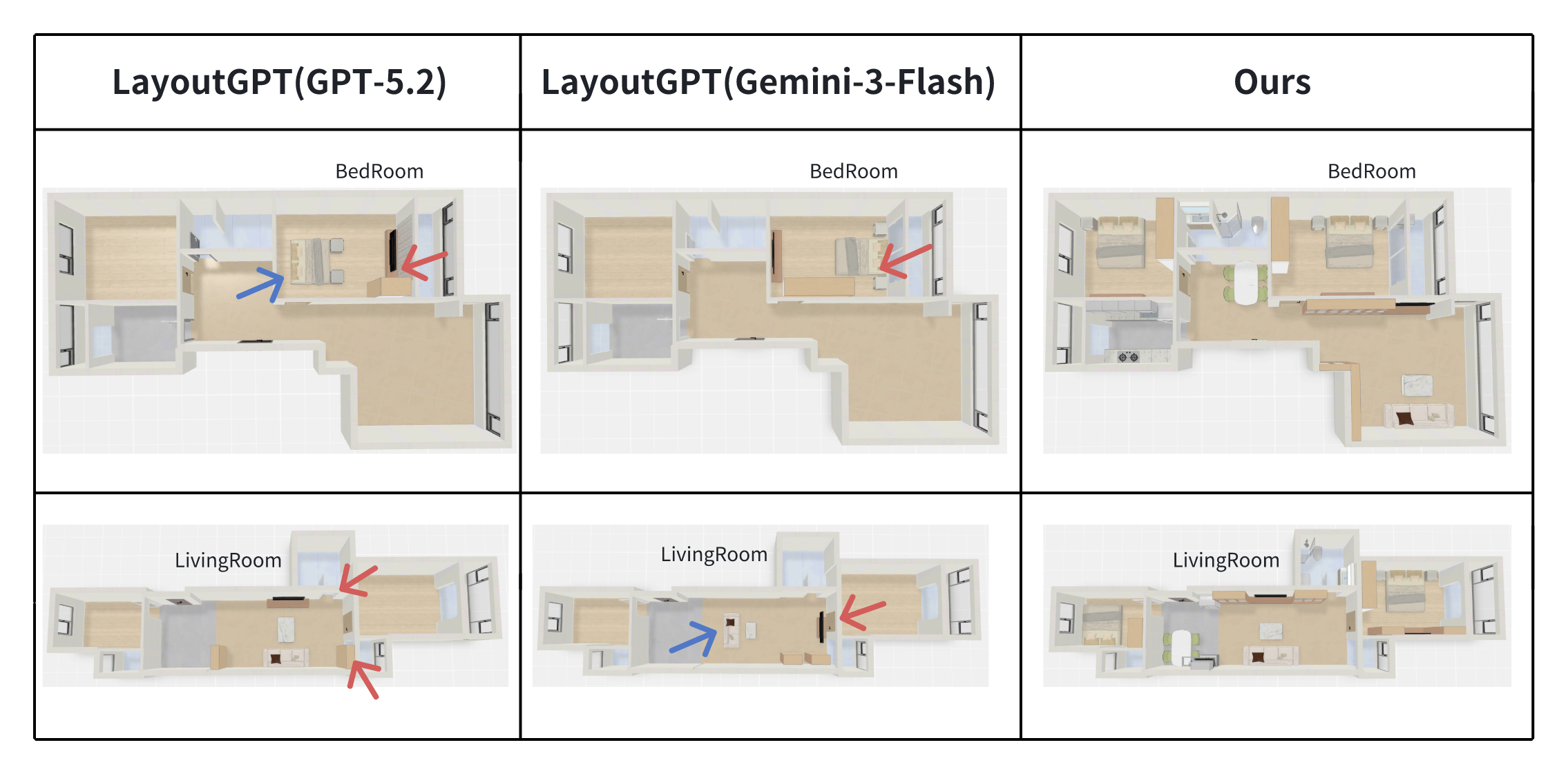}
    \caption{Visual comparison with LayoutGPT. Red arrows indicate furniture accessibility issues, and blue arrows indicate furniture misalignment with walls. RenoLayout produces more accessible and better-aligned room layouts in these examples.}
    \label{fig:layoutgpt}
\end{figure}
\section{Conclusion}

We introduced \textbf{RenoLayout}, a unified framework for constraint-aware residential design. By combining the Design Map, difference-grounded design-point annotation, Geometry-as-Code representation, and CPT--SFT--RL training, RenoLayout jointly improves instruction following, constraint compliance, and living quality. The final RenoLayout achieves 91.63\% Recall@1 and 61.00\% Pass@1, demonstrating the effectiveness of combining preference optimization with multi-rule-guided reinforcement learning.

\clearpage
\bibliography{aaai2027}

\clearpage

\setcounter{secnumdepth}{1}

\appendix
\renewcommand{\thesection}{Appendix~\Alph{section}}

\section{Training Data and Data Mixture}
\label{appendix:data_composition}

\noindent\textbf{Data Sources and Usage. }
The training data used across CPT, SFT, and RL are derived from real-world residential floorplans. All records are anonymized to remove personally identifiable information as well as location-related metadata. These training corpora cannot be publicly released because they are derived from proprietary commercial data governed by data-use agreements and are subject to residential privacy constraints. Existing public alternatives do not provide paired before--after residential designs with complete furnishing. They also lack the design-critical structural and building-service annotations required for training RenoLayout. The RenoLayout Benchmark contains 200 quality-filtered paired real-world residential-design cases for reproducible evaluation and will be publicly released through the OpenReview submission materials. All benchmark cases are strictly disjoint from the data used for CPT, SFT, and RL.

The SFT corpus contains approximately 34,000 real-world floorplans before augmentation. We apply rotational augmentation to the paired floorplans, yielding approximately 130,000 SFT training samples (0.7 billion tokens).

We serialize floorplan vectors as executable Python code so that spatial dependencies, such as room adjacency and wall closure, are represented through explicit program structure. CPT uses approximately 40 billion tokens partitioned into the subsets summarized in Table~\ref{tab:data_composition}.

\noindent\textbf{Data Mixing Strategy}
To mitigate catastrophic forgetting \cite{Code_llama}, we mix general-purpose data with domain-specific vector code. The four components serve complementary roles:
\begin{center}
\centering
\scriptsize
\setlength{\tabcolsep}{2pt}
\renewcommand{\arraystretch}{1.0}
\begin{tabularx}{\columnwidth}{@{}>{\raggedright\arraybackslash}p{1.50cm} >{\centering\arraybackslash}p{0.65cm} X@{}}
\toprule
\textbf{Component} & \textbf{Ratio}  & \textbf{Rationale} \\
\midrule
\textbf{Vector Code} & 90\%  & Core \textit{Geometry-as-Code} generation tasks for spatial generation and interpretation. \\
\textbf{General Python} \cite{fujii2025rewritingpretrainingdataboosts} & 5\% & Prevents syntax degradation and maintains code-structure fluency. \\
\textbf{Mathematical Data} \cite{fujii2025rewritingpretrainingdataboosts} & 4\%  & Enhances logical reasoning and numerical precision for coordinate arithmetic. \\
\textbf{Domain NL (from the internet)} & 1\%  & Ensures robust alignment between user instructions and geometric execution. \\
\bottomrule
\end{tabularx}
\captionof{table}{Composition of the CPT Training Corpus.}
\label{tab:data_composition}
\end{center}

\section{Experimental Setup and Reproducibility}
\label{appendix:experimental_setup}

All data shuffling, training, and generation used a fixed random seed of 42. We retained one fixed hyperparameter configuration per training stage because training was stable. All stages used the AdamW optimizer with a weight decay of 0.01 under bfloat16 (BF16) precision. SFT used a warmup ratio of 0.1. RL sampling used a temperature of 0.6; evaluation-time inference used a temperature of 0.3 and top-$p$ of 0.8. For each evaluated model configuration, we independently sampled each of the 200 benchmark cases 10 times, yielding 2,000 generations per configuration; consistent with their definitions, the reported Recall@1 and Pass@1 evaluate the first sample for each case. Hardware varied by stage: 8--16 NVIDIA H200 GPUs, 180 Intel Xeon Platinum 8558 CPU cores, and 1,980 GB of system memory. The software environment was Debian-based GNU/Linux (kernel 5.4.250-9, x86\_64), Python 3.10, PyTorch 2.4.0 with CUDA 12.4, and Transformers 4.51.3.


\section{Complete Design Map}
\label{appendix:design_map}

The complete Design Map contains 123 fine-grained design points, comprising 52 structural-renovation design points and 71 furniture-layout design points. The full taxonomy is listed below using the names employed throughout the paper and evaluation pipeline.

\begin{table*}[!t]
\centering
\scriptsize
\setlength{\tabcolsep}{2.2pt}
\renewcommand{\arraystretch}{0.92}
\begin{tabularx}{\textwidth}{@{}r>{\raggedright\arraybackslash}X r>{\raggedright\arraybackslash}X r>{\raggedright\arraybackslash}X@{}}
\toprule
\multicolumn{6}{c}{\textbf{Structural-Renovation Design Points (52)}} \\
\midrule
\textbf{No.} & \textbf{Design Point} & \textbf{No.} & \textbf{Design Point} & \textbf{No.} & \textbf{Design Point} \\
\midrule
1 & Bedroom Storage & 19 & Remove Entryway & 36 & 4BR to 3BR \\
2 & Door/Window Adjustment & 20 & 2BR to 3BR & 37 & Master Suite \\
3 & Wet-Dry Separation (2-Zone) & 21 & Add Study in Living Room & 38 & Wet-Dry Separation (3-Zone) \\
4 & Convert to Utility Area & 22 & Enlarge Bedroom & 39 & Kitchen-Dining Integration \\
5 & Entryway Storage & 23 & 3BR to 2BR & 40 & Corridor Storage \\
6 & Entryway/Corridor Space Reallocation & 24 & Shrink Kitchen & 41 & Storage Room Organization \\
7 & Convert to Open Kitchen & 25 & Enlarge Entryway & 42 & 3BR to 4BR \\
8 & LDK Integration & 26 & 1BR to 2BR & 43 & Study \& Storage Room \\
9 & Merge Balcony into Adjacent Room & 27 & Entryway Renovation & 44 & Living-Dining Integration \\
10 & Convert to Semi-Open Kitchen & 28 & Merge Kitchen \& Balcony & 45 & Allocate Corridor to Bedroom \\
11 & Enlarge Bathroom & 29 & Living Room Storage & 46 & Allocate Corridor to Bathroom \\
12 & Shared Wall Zigzag Remodel & 30 & 2BR to 1BR & 47 & Convert to Leisure Space \\
13 & Dining Room Storage & 31 & Multi-Zone Utilization & 48 & Convert Storage to Study \\
14 & Add Ensuite to Bedroom & 32 & Add Storage Room & 49 & Chinese-Western Kitchen Zoning \\
15 & Convert Bedroom to Utility Room & 33 & Add Walk-in Closet to Bedroom & 50 & Convert to Storage Space \\
16 & Enlarge Kitchen & 34 & Shrink Entryway & 51 & Wet-Dry Separation (4-Zone) \\
17 & Incorporate into Other Space & 35 & Add Closet \& Ensuite to Bedroom & 52 & Merge into Other Space \\
18 & Balcony Storage & & & & \\
\midrule
\multicolumn{6}{c}{\textbf{Furniture-Layout Design Points (71)}} \\
\midrule
\textbf{No.} & \textbf{Design Point} & \textbf{No.} & \textbf{Design Point} & \textbf{No.} & \textbf{Design Point} \\
\midrule
1 & Bathroom Vanity - Built-in Sink & 25 & Kitchen Cabinet - U-Shaped & 49 & Living Room - Bar Counter Divider \\
2 & Bathroom Vanity - Freestanding Cabinet & 26 & Bedroom Bed - Tatami w/ Side Desk & 50 & Wide Hall - Sofa as Center Divider \\
3 & Standard Set - Bed, Nightstand, Wardrobe & 27 & Wide Hall - Parallel Layout & 51 & Standard Layout - Full-Wall Bookshelf \\
4 & Living Room - Corner Tea Table & 28 & Wide Hall - Sofa \& Table Parallel & 52 & Kitchen Bar - Bar Extends Countertop \\
5 & Kitchen Cabinet - L-Shaped & 29 & Kitchen Cabinet - Galley & 53 & Bathroom Toilet - Sunken Squat Toilet \\
6 & Kitchen Workflow - Golden Triangle & 30 & Bedroom Closet - Wardrobe Partition & 54 & Kitchen Island - Freestanding \\
7 & Kitchen Appliance - Fridge Near Entrance & 31 & Bedroom Convert - Desk \& Bookshelf & 55 & Study Room - Side-by-Side Double Desk \\
8 & Fridge Placement - In Kitchen & 32 & Bedroom Bed - Platform w/ Drawers & 56 & Living Room Reading - Desk w/ Bookshelf \\
9 & Bedroom Storage - Full-Wall Wardrobe & 33 & Wide Hall - Vertical Layout & 57 & Dining Storage - Low Sideboard \\
10 & Fridge Placement - In Living Room & 34 & Bedroom Office - Window Long Desk (2P) & 58 & Wide Hall - Sofa Back w/ Bar Divider \\
11 & Washer Placement - In Bathroom & 35 & Fridge Placement - In Dining Room & 59 & Balcony Zoning - Utility + Leisure \\
12 & Washer Placement - Balcony w/ Counter & 36 & Living Room - Wall Bookshelf \& Float TV & 60 & Dining Zone - Island as Dining Table \\
13 & Balcony Utility - Washer \& Drying Rack & 37 & Dining Storage - Full-Height Sideboard & 61 & Corridor Storage - Built-in Cabinet \\
14 & Bedroom Office - Desk by Window & 38 & Dining Table - Rectangular & 62 & Standard Set - Storage Cabinet \\
15 & Kitchen Appliance - Washer Under Counter & 39 & Study Room - Desk \& Bookshelf & 63 & Dining Banquette - Wall-Side Bench \\
16 & Living Room - Remove TV Stand & 40 & Bathroom Toilet - Wall-Hung w/ Concealed Tank & 64 & Bathroom Vanity - Wall-Mounted Sink \\
17 & Kitchen Cabinet - I-Shaped & 41 & Living Room Office - Desk Behind Sofa & 65 & Balcony Leisure - Tea Table \& Chair \\
18 & Entryway - Standard Shoe Cabinet & 42 & Wide Hall LDK - Island Divides L/D & 66 & Entryway - Full-Height Coat Closet \\
19 & Washer Placement - In Kitchen Cabinet & 43 & Standard Layout - Centered Desk & 67 & Living Room - TV Wall Center Divider \\
20 & Bathroom Vanity - External Sink & 44 & Bathroom Bathtub - Built-in Wall-Side & 68 & Living Room Reading - Island Desk \\
21 & Bathroom Shower - Rectangular Wall-Mount & 45 & Bathroom Shower - Linear Glass Partition & 69 & Study/Guest Room - Sofa Bed \& Desk \\
22 & Living Room - Sofa as Divider & 46 & Living Room - Projector Replaces TV & 70 & Living Room Reading - Desk \& Chair \\
23 & Kitchen Built-in - Tall Cabinet w/ Fridge & 47 & Entryway - Slim Flip Shoe Cabinet & 71 & Dining Table - Round \\
24 & Bathroom Shower - Diamond Corner Unit & 48 & Dining Zone - Table Center \& Sideboard & & \\
\bottomrule
\end{tabularx}
\caption{Complete Design Map containing 52 structural-renovation and 71 furniture-layout design points. Entries are arranged in three parallel numbered blocks for compact presentation.}
\label{tab:complete_design_map}
\end{table*}

\section{Evaluation Metrics}
\label{metrics}
Table~\ref{tab:evaluation_metrics} summarizes the evaluation framework. Recall@1 measures design-point fulfillment; Pass@1 requires all 22 constraint-compliance checks to pass; and VLM-HS combines five subjective dimensions with a sixth physical-compliance dimension. PAS provides an overall professional comparison between RenoLayout outputs and corresponding ground-truth designs.

\begin{table*}[!t]
  \centering
  \scriptsize
  \renewcommand{\arraystretch}{1.05}
  \begin{tabularx}{\textwidth}{>{\raggedright\arraybackslash}p{2.6cm} >{\raggedright\arraybackslash}p{4.1cm} >{\centering\arraybackslash}p{1.5cm} X}
    \toprule
    \textbf{Dimension} & \textbf{Metric} & \textbf{Range} & \textbf{Description and Key Indicators} \\
    \midrule
    Design-point fulfillment & Recall@1 & 0--100\% & Mean case-level proportion of requested design points fulfilled in the first generated outputs, averaged across all 200 cases by a rule-based evaluator. \\
    Constraint compliance & Pass@1 & 0--100\% & Percentage of one-sample generated results that pass all 22 checks across geometric validity, structural compliance, functional integrity, and collision-free placement. \\
    Constraint compliance & Geometric Validity ($\mathcal{P}_{\mathrm{geom}}$) & 0--100\% & Percentage of generated designs with well-formed polygons, valid references, and complete required fields. \\
    Constraint compliance & Structural Compliance ($\mathcal{P}_{\mathrm{struct}}$) & 0--100\% & Percentage of generated designs that preserve load-bearing walls, exterior envelopes, and shafts while maintaining room connectivity through valid doors. \\
    Constraint compliance & Functional Integrity ($\mathcal{P}_{\mathrm{func}}$) & 0--100\% & Percentage of generated designs that retain fixed infrastructure and essential fixtures without damage or omission. \\
    Constraint compliance & Collision-Free Placement ($\mathcal{P}_{\mathrm{coll}}$) & 0--100\% & Percentage of generated designs without furniture overlap or forbidden-zone violations. \\
Living quality & VLM-based Habitability Score (VLM-HS) & 0--10 & Four separately reported model-specific scores, each combining five weighted subjective dimensions with a fused physical-compliance score and offline ground-truth deficit calibration. \\
Professional approval & Professional Approval Score (PAS) & 0--10 & Equally weighted mean of eight professional-design dimensions, averaged over all 200 cases; reported for RenoLayout outputs and corresponding ground-truth designs. \\
    \bottomrule
  \end{tabularx}
\caption{Overview of the unified evaluation framework.}
\label{tab:evaluation_metrics}
\end{table*}

\noindent\textbf{Professional Approval Score (PAS).}
For each case, each professional designer independently assigned eight 0--10 scores covering Structural Integrity \& Compliance, Connectivity \& Accessibility, Functional Layout, Spatial Layout, Functional Zoning, Circulation Quality, Amenity Configuration, and Layout Configuration. The case-level PAS is their equally weighted mean:
\begin{equation}
\mathrm{PAS}_{i,d}=\frac{1}{8}\sum_{k=1}^{8}s_{i,d,k},
\end{equation}
where $s_{i,d,k}$ is the score assigned to case $i$ by designer $d$ for dimension $k$. All 200 benchmark cases were included without filtering. Each designer-level PAS is the mean over the 200 cases. The reported overall Mean and Std. are computed across the three designer-level scores.

Each constraint-compliance check produces a binary violation flag. A submetric passes only when none of its constituent flags are triggered, and Pass@1 requires all 22 checks across the four submetrics to pass. Table~\ref{tab:submetric_checks} reports the exact rule composition used by the evaluator.

\begin{table*}[!t]
  \centering
  \scriptsize
  \setlength{\tabcolsep}{4pt}
  \renewcommand{\arraystretch}{1.08}
  \begin{tabularx}{\textwidth}{@{}>{\raggedright\arraybackslash}p{2.2cm} >{\centering\arraybackslash}p{1.2cm} >{\raggedright\arraybackslash}X@{}}
    \toprule
    \textbf{Submetric} & \textbf{Checks} & \textbf{Violation Flags and Annotations} \\
    \midrule
    Geometric Validity ($\mathcal{P}_{\mathrm{geom}}$) & 5 &
    \texttt{polygon\_invalid}: malformed polygon geometry;\newline
    \texttt{polygon\_self\_intersect}: polygon edges intersect one another;\newline
    \texttt{invalid\_reference}: an entity references a nonexistent or invalid object;\newline
    \texttt{missing\_required\_field}: a mandatory geometry or semantic field is absent;\newline
    \texttt{wall\_endpoint\_in\_opening}: a wall endpoint lies inside a door or window opening \\
    Structural Compliance ($\mathcal{P}_{\mathrm{struct}}$) & 11 &
    \texttt{load\_bearing\_wall\_removed}: an existing load-bearing wall is deleted;\newline
    \texttt{load\_bearing\_wall\_moved}: a load-bearing wall is displaced;\newline
    \texttt{load\_bearing\_wall\_thinned}: load-bearing wall thickness is reduced;\newline
    \texttt{load\_bearing\_wall\_new\_opening}: a new opening is cut into a load-bearing wall;\newline
    \texttt{load\_bearing\_wall\_opening\_enlarged}: an existing load-bearing-wall opening is enlarged;\newline
    \texttt{exterior\_wall\_altered}: the exterior envelope geometry is modified;\newline
    \texttt{facade\_new\_window}: a new window is introduced on the facade;\newline
    \texttt{wall\_crosses\_shaft}: a wall intersects a protected service shaft;\newline
    \texttt{opening\_exceeds\_wall}: an opening extends beyond its supporting wall segment;\newline
    \texttt{no\_entry\_door}: the dwelling has no valid entrance door;\newline
    \texttt{room\_no\_door}: a room has no valid doorway \\
    Functional Integrity ($\mathcal{P}_{\mathrm{func}}$) & 4 &
    \texttt{kitchen\_flue\_altered}: the fixed kitchen exhaust flue is modified;\newline
    \texttt{gas\_pipe\_altered}: the fixed gas-pipe location or geometry is modified;\newline
    \texttt{entry\_door\_altered}: the fixed entrance door is modified;\newline
    \texttt{bathroom\_missing\_fixture}: an essential bathroom fixture is absent \\
    Collision-Free Placement ($\mathcal{P}_{\mathrm{coll}}$) & 2 &
    \texttt{item\_overlap}: furniture or other placed items overlap;\newline
    \texttt{furniture\_forbidden\_position}: furniture occupies a prohibited wall, opening, or clearance zone \\
    \midrule
    \textbf{Total} & \textbf{22} & \textbf{Pass@1 requires that none of the 22 violation flags be triggered.} \\
    \bottomrule
  \end{tabularx}
  \caption{Exact rule composition of the four constraint-compliance submetrics, with a brief annotation for each violation flag.}
  \label{tab:submetric_checks}
\end{table*}

\section{Annotation Auditing}
\label{appendix:annotation_auditing}

We use VLMs for two distinct purposes. First, multiple VLMs, including Seed, GPT, and Gemini, serve as annotation auditors. Given the pre-design image, post-design image, extracted before--after differences, and rule-derived design-point labels, they identify unsupported, conflicting, ambiguous, or duplicated labels. They do not generate labels, and ambiguous cases are subjected to manual inspection.

A four-VLM panel comprises Seed2.1 Pro~\cite{doubao_seed21}, GLM-5V-Turbo~\cite{glm5team2026glm5vibecodingagentic}, GPT-5.5~\cite{gpt55}, and Gemini-3.1-Flash-Lite~\cite{gemini31flashlite}. This panel is employed for VLM-HS evaluation. Appendix~\ref{appendix:vlm_scoring} details the inputs, scoring dimensions, aggregation scheme, and complete prompts.

\section{Failure Case Analysis}
\begin{figure*}[t]
    \centering
    \includegraphics[width=0.9\linewidth]{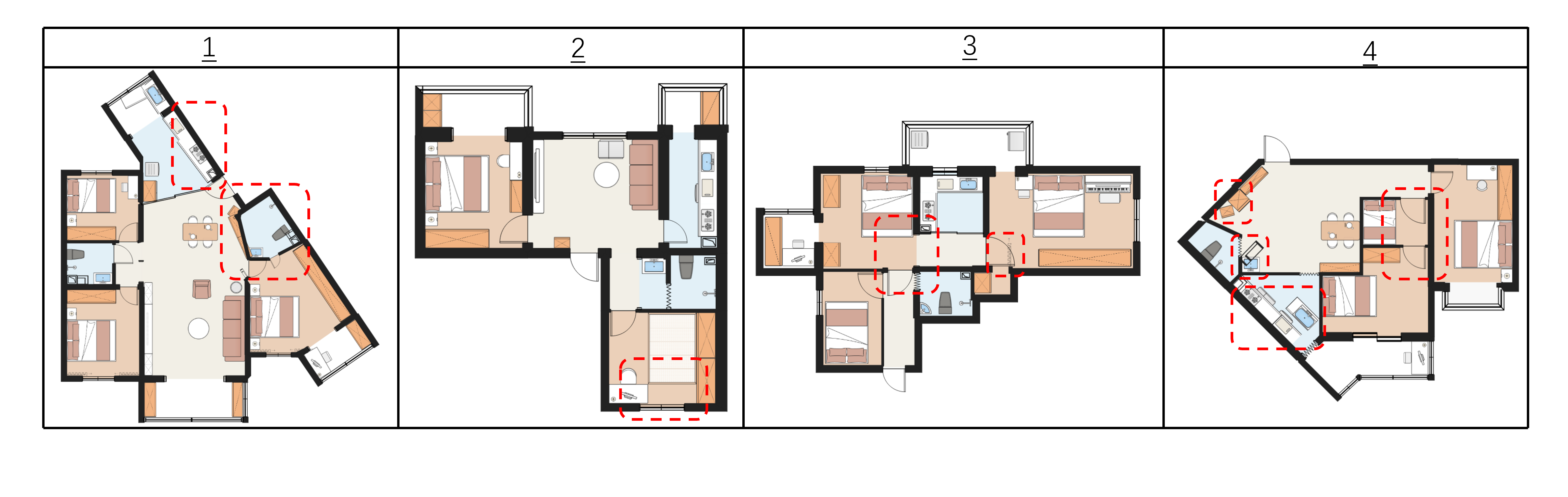}
    \caption{Representative failure cases involving irregular boundaries and severely constrained spaces.}
    \label{fig:failcase}
\end{figure*}

We visualize typical failure cases in Figure~\ref{fig:failcase}.
The primary errors occur when dealing with \textbf{complex boundaries}.
As seen in Case 1 and Case 4, when the input outline is highly irregular, the model struggles to partition the space effectively, leading to disconnected rooms or invalid overlaps.
Another common failure arises in \textbf{extremely tight spaces}.
In Case 2 and Case 3, the limited area forces the model to generate layouts where furniture obstructs doors or hallways become too narrow.
These examples indicate that while our method performs well on standard layouts, it still faces challenges when handling extreme geometric irregularities or severe spatial constraints.

\section{AI Assistant Usage Disclosure}

\noindent\textbf{Research and Data Annotation.}
We used multiple vision-language models, including Seed, GPT, and Gemini, to audit rule-derived design-point annotations against observable differences between paired before--after designs. These auditors did not generate the ground-truth labels; the retained labels remain grounded in explicit before--after transformations, with ambiguous cases subjected to manual inspection. Separately, a fixed four-VLM panel (comprising Seed2.1 Pro, GLM-5V-Turbo, GPT-5.5, and Gemini-3.1-Flash-Lite) was employed for VLM-HS evaluation.

\noindent\textbf{Coding.}
AI coding assistants (e.g., GitHub Copilot) were used to assist in writing the infrastructure code for the evaluation pipeline. All AI-suggested code was thoroughly tested and verified by the research team.

\noindent\textbf{Writing and Polishing.}
We used ChatGPT/Gemini for grammar correction, stylistic polishing, and rephrasing. This assistance improved manuscript clarity as well as flow. Human authors entirely conceived the core arguments, technical descriptions, and data interpretations. They also wrote the manuscript.

\section{VLM-Based Scoring Protocol}
\label{appendix:vlm_scoring}

This appendix details the VLM-based scoring protocol used for the Habitability Scores reported in Appendix~\ref{metrics}. We use four independently evaluated VLMs: Seed2.1 Pro, GLM-5V-Turbo, GPT-5.5, and Gemini-3.1-Flash-Lite. Each model assesses the rendered floor plan together with the design-point specification and deterministic physical-inspection facts. A deterministic geometric engine provides an objective physical signal. For each VLM separately, its score is calibrated offline against the independently evaluated ground-truth plan for the same case. We retain and report the four model-specific VLM-HS metrics separately.

\subsection{Evaluation Dimensions}

Table~\ref{tab:vlm_dims} summarizes the six dimensions. Each VLM returns an integer score in $[0,10]$ for every dimension, where higher is better. The first five dimensions form the abstract design-quality score. The sixth dimension provides the VLM's holistic physical judgment, which is fused with the deterministic engine score to form the final physical score.

\begin{center}
\centering
\small
\begin{tabularx}{\columnwidth}{@{}c>{\raggedright\arraybackslash}X>{\raggedright\arraybackslash}p{1.55cm}>{\centering\arraybackslash}p{1.20cm}@{}}
\toprule
\textbf{No.} & \textbf{Dimension} & \textbf{Type} & \textbf{Weight $w_i$} \\
\midrule
1 & Functional Zoning & Abstract & 1.42 \\
2 & Spatial Layout & Abstract & 1.01 \\
3 & Circulation Quality & Abstract & 1.26 \\
4 & Amenity Configuration & Abstract & 1.00 \\
5 & Design-Point Implementation & Abstract & 1.20 \\
6 & Physical Compliance & VLM physical review & --- (fusion) \\
\bottomrule
\end{tabularx}
\captionof{table}{Evaluation dimensions and aggregation weights.}
\label{tab:vlm_dims}
\end{center}

\subsection{Physical Inspection Layer}
A local geometric engine inspects the structured floor-plan output and assigns each detected issue one of three severity levels, as summarized in Table~\ref{tab:vlm_phys}. P0 and P1 are used in the engine-calibrated physical score; P2 is retained for diagnosis but does not directly reduce that score.

\begin{center}
\centering
\small
\begin{tabularx}{\columnwidth}{@{}>{\raggedright\arraybackslash}p{1.35cm} >{\raggedright\arraybackslash}p{2.25cm} >{\raggedright\arraybackslash}X@{}}
\toprule
\textbf{Level} & \textbf{Meaning} & \textbf{Representative issues} \\
\midrule
P0 & Critical; unsafe, unusable, or non-compliant & Invalid structural modification, room overlap, missing essential functions, inaccessible space, or severe wall/opening conflicts \\
P1 & Severe; substantially degrades normal use & Insufficient clearance, narrow circulation, door blockage, furniture overlap, inadequate room area, or functional adjacency conflicts \\
P2 & Advisory; minor optimization issue & Local dimensional, lighting, storage, or configuration suggestions \\
\bottomrule
\end{tabularx}
\captionof{table}{Severity levels of physical defects.}
\label{tab:vlm_phys}
\end{center}

\subsection{Score Aggregation}

\noindent\textbf{Model-specific dimensional scores.}
Let $s_i^{(m)}$ denote the score assigned to dimension $i$ by VLM $m$, where $m=1,\dots,M$ and $M=4$. All subsequent quantities are computed independently for each model $m$.

\noindent\textbf{Abstract design-quality score.}
For model $m$, the first five dimensions are aggregated using the weights in Table~\ref{tab:vlm_dims}:
\begin{equation}
S_a^{(m)}=
\frac{\sum_{i=1}^{5}w_i\,s_i^{(m)}}
{\sum_{i=1}^{5}w_i}.
\end{equation}

\noindent\textbf{Engine-calibrated physical score.}
Let $n_{P0}$, $n_{P1}$, and $n_{P2}$ be the numbers of engine-detected issues at the three severity levels. The deterministic physical score, shared across the four VLMs for a given plan, is
\begin{equation}
S_e=
\operatorname{clip}
\left(
10-1.25n_{P0}-0.06n_{P1}-0.00n_{P2},
0,10
\right).
\end{equation}
This formulation makes P0 violations the dominant physical ordering signal, uses P1 violations for finer within-band differentiation, and retains P2 only as diagnostic evidence.

\noindent\textbf{Model-specific fused physical score.}
For model $m$, its sixth-dimensional score $S_v^{(m)}=s_6^{(m)}$ provides a holistic visual judgment of physical usability. We combine it with the engine score as
\begin{equation}
\tilde{S}_v^{(m)}=\min(S_v^{(m)},S_e),
\end{equation}
\begin{equation}
S_p^{(m)}=0.70S_e+0.30\tilde{S}_v^{(m)}.
\end{equation}
The minimum operator ensures that a VLM judgment cannot offset deterministic P0/P1 violations. When a VLM identifies additional visual usability concerns and assigns a lower score than the engine, it contributes an additional penalty through the fused score.

\noindent\textbf{Model-specific absolute habitability score.}
Before reference calibration, the score for model $m$ is
\begin{equation}
S_{\mathrm{raw}}^{(m)}=
0.45S_a^{(m)}+0.55S_p^{(m)}.
\end{equation}

\noindent\textbf{Offline ground-truth deficit calibration.}
For each test case, the predicted plan and its ground-truth plan are evaluated independently by the same VLM $m$. Let $S_a^{\mathrm{gt},(m)}$ denote the ground-truth abstract score, whereas $S_p^{\mathrm{gt},(m)}$ denotes the fused physical score. We compute a one-sided reference deficit:
\begin{equation}
\begin{aligned}
D_{\mathrm{gt}}^{(m)}={}&0.45\max(0,S_a^{\mathrm{gt},(m)}-S_a^{(m)}) \\
&+0.55\max(0,S_p^{\mathrm{gt},(m)}-S_p^{(m)}).
\end{aligned}
\end{equation}
Only deficits relative to the ground truth are penalized; a prediction does not receive an additional bonus for exceeding the ground truth on an individual component.

\noindent\textbf{Final model-specific VLM-HS.}
The final score produced by model $m$ is
\begin{equation}
S_u^{(m)}=
\operatorname{clip}
\left(
S_{\mathrm{raw}}^{(m)}-\gamma D_{\mathrm{gt}}^{(m)},
0,10
\right),
\end{equation}
where $\gamma=1.0$ by default. For a ground-truth plan evaluated against itself, $D_{\mathrm{gt}}^{(m)}=0$. For an evaluation set of $N$ cases, we report the four model-specific metrics
\begin{equation}
\mathrm{VLM\text{-}HS}^{(m)}=
\frac{1}{N}\sum_{j=1}^{N}S_{u,j}^{(m)},
\qquad m=1,\dots,4.
\end{equation}

\subsection{Scoring Prompt}

Tables~\ref{tab:vlm_prompt_1}--\ref{tab:vlm_prompt_5} reproduce the scoring prompt in component order. We submit the same prompt independently to Seed2.1 Pro, GLM-5V-Turbo, GPT-5.5, and Gemini-3.1-Flash-Lite. The resulting scores are reported as model-specific metrics. Each VLM receives the post-design rendering, the pre-design rendering when available, the requested design points, and deterministic P0/P1 physical-inspection facts. Ground-truth plans are not included in the VLM inputs; ground-truth deficit calibration is computed after independent scoring. Dimension names follow Table~\ref{tab:vlm_dims}.

\begin{table*}[p]
\centering
\scriptsize
\renewcommand{\arraystretch}{1.08}
\begin{tabularx}{\textwidth}{@{}>{\raggedright\arraybackslash}p{3.4cm}>{\raggedright\arraybackslash}X@{}}
\toprule
\textbf{Prompt component} & \textbf{Instruction content} \\
\midrule
Role and task & You are a senior residential design reviewer. Evaluate the post-design plan dimension by dimension. Assess subjective design quality that geometry cannot fully determine in Dimensions 1--5. Assess physical compliance together with hard defects only in Dimension 6. \\
\midrule
Design context and design points & When both images are available, they depict the same dwelling in two states. The first image shows the original plan, whereas the second presents the post-design plan under review. Score only the post-design plan; the original plan is provided solely to explain what changed and why. All scores reflect the absolute quality of the post-design solution, not improvement relative to the original plan. The listed design points are the required design objectives. Dimensions 1--4 evaluate absolute zoning, layout, circulation, and amenity quality. They must not mechanically reward or penalize compliance with a design point. However, consequences necessarily caused by correctly implementing a required design point must not be treated as defects. Dimension 5 alone evaluates whether design points were implemented. It also assesses whether their implementation is natural and coherent. \\
\midrule
Objective physical-inspection facts & [Objective physical-inspection facts detected by the deterministic engine; the primary basis for Dimension 6.] \newline
\{Insert the engine's P0/P1 fact list here; if there are no hits, write: No P0/P1-level physical issues detected.\} \newline
Cross-check the list against the post-design image. Clearly inconsistent entries may be treated as false detections and downweighted. Do not fabricate physical issues beyond the list. The fact list is used only for Dimension 6 and must not adjust Dimensions 1--5. \\
\bottomrule
\end{tabularx}
\caption{VLM-HS scoring prompt, Part I: role, design context, and physical-inspection input.}
\label{tab:vlm_prompt_1}
\end{table*}

\begin{table*}[p]
\centering
\scriptsize
\renewcommand{\arraystretch}{1.08}
\begin{tabularx}{\textwidth}{@{}>{\raggedright\arraybackslash}p{3.4cm}>{\raggedright\arraybackslash}X@{}}
\toprule
\textbf{Prompt component} & \textbf{Instruction content} \\
\midrule
Dimension 1: Functional Zoning & Evaluate whether active/quiet, public/private, and wet/dry zones are clearly layered with limited mutual interference. Check whether bedrooms form an independent cluster separated from living/dining/kitchen spaces. Assess whether the plan progresses from entry to public/social space and then reaches private sleeping space. Determine whether bathrooms serve either the sleeping area or the public area rather than remaining isolated centrally. Finally, assess whether living, dining, and kitchen areas form a coherent social zone. This dimension evaluates zoning logic, not missing furniture or reachability. \\
\midrule
Dimension 2: Spatial Layout & Evaluate whether proportions are comfortable and balanced. Determine whether the circulation area is economical and the primary/secondary hierarchy is appropriate. Check whether major rooms are roughly within a 1:1--1:2 aspect-ratio range. Assess whether corridors are compact rather than long, dark, or wasteful. Determine whether the master bedroom is larger than secondary bedrooms. The living room should be the largest or second-largest public space. Finally, check whether obvious fragmentary space exists. This dimension evaluates proportion and layout rather than exact floor area. \\
\midrule
Dimension 3: Circulation Quality & Assuming rooms are reachable with no must-pass-through-bedroom defect, evaluate the smoothness and quality of daily circulation. Check whether the entry provides a foyer or dust-buffer zone. Assess whether kitchen and dining areas are adjacent with a short housework path. Determine whether the nighttime bedroom-to-bathroom path is short, quiet, and minimally exposed to public activity. This dimension evaluates circulation quality, not objective reachability. \\
\bottomrule
\end{tabularx}
\caption{VLM-HS scoring prompt, Part II: definitions of Dimensions 1--3.}
\label{tab:vlm_prompt_2}
\end{table*}

\begin{table*}[p]
\centering
\scriptsize
\renewcommand{\arraystretch}{1.08}
\begin{tabularx}{\textwidth}{@{}>{\raggedright\arraybackslash}p{3.4cm}>{\raggedright\arraybackslash}X@{}}
\toprule
\textbf{Prompt component} & \textbf{Instruction content} \\
\midrule
Dimension 4: Amenity Configuration & Evaluate whether the plan provides hallmark configurations of quality living. Check for a master suite with a dedicated bathroom or dressing/storage area. Also assess whether the plan includes a distinct entry foyer with systematic storage. Where multiple bathrooms exist, evaluate owner/guest separation. Finally, assess whether an open, connected living/dining area supports social use. Evaluate only items visible in the image; skip unclear evidence rather than fabricating it. \\
\midrule
Dimension 5: Design-Point Implementation & Evaluate the listed design points in terms of both fulfillment and implementation quality. Fulfillment asks whether each required design point is visibly realized without omission, distortion, or reversal. Implementation quality considers location, scale, and spatial relationships. These aspects should appear natural and coherent with the overall plan rather than forced or achieved at the expense of visible living quality. This dimension must be supported by a mandatory item-by-item design-point audit. \\
\midrule
Dimension 6: Physical Compliance & Evaluate objective physical compliance and defects. Use the deterministic P0/P1 fact list as the primary basis, cross-checked against the post-design image. P0 defects are the primary ordering signal and must be strongly penalized; P1 defects provide only finer differentiation within a severity band. Do not invent issues beyond the fact list. \\
\midrule
Scope boundary & The following objective items are evaluated only in Dimension 6: room reachability; furniture overlap, through-wall placement, or door blockage; missing essential furniture; natural daylight; must-pass-through-bedroom circulation; direct kitchen-to-toilet-bathroom connection; door or passage width; and load-bearing/exterior-wall modifications. Dimensions 1--5 evaluate only subjective design quality within their stated scope. Design-point implementation is evaluated primarily in Dimension 5, while correctly required design-point consequences must not be mechanically treated as defects in Dimensions 1--4. \\
\bottomrule
\end{tabularx}
\caption{VLM-HS scoring prompt, Part III: definitions of Dimensions 4--6 and scope boundaries.}
\label{tab:vlm_prompt_3}
\end{table*}

\begin{table*}[p]
\centering
\scriptsize
\renewcommand{\arraystretch}{1.08}
\begin{tabularx}{\textwidth}{@{}>{\raggedright\arraybackslash}p{3.4cm}>{\raggedright\arraybackslash}X@{}}
\toprule
\textbf{Prompt component} & \textbf{Instruction content} \\
\midrule
Deduction/bonus checklist & \textbf{Functional Zoning}\quad Deductions: public functions interrupted by bedrooms or bathrooms; rooms arranged as a single-axis sequence without clustering; entry directly entering or facing the bedroom zone; a bedroom sandwiched between public spaces; or a centrally isolated bathroom. Bonuses: bedrooms grouped independently behind a corridor or foyer, and a clear entry $\rightarrow$ public $\rightarrow$ private progression.\newline
\textbf{Circulation Quality}\quad Deductions: no foyer buffer; kitchen and dining separated by a detour; nighttime bedroom-to-bathroom circulation crossing public activity space; crossing or strongly interfering daily paths; or an obviously circuitous route from entry to major functions. Bonuses: an entry buffer, adjacent kitchen/dining spaces, and a short private nighttime path.\newline
\textbf{Amenity Configuration}\quad Deductions: an ordinary master bedroom with neither dedicated bathroom nor dressing/storage area; no entry zone or shoe-storage location; scattered storage; multiple bathrooms without owner/guest separation; or a living/dining area fragmented by circulation. Bonuses: a master suite, foyer plus systematic storage, owner/guest-separated bathrooms, and an open connected living/dining area.\newline
\textbf{Design-Point Implementation}\quad Audit every listed design point individually. A point is \emph{Implemented} only when its required action produces the intended outcome. \emph{Partially implemented} indicates that the direction is present but a key action, scale, or placement is missing. \emph{Not implemented} indicates no visible evidence. \emph{Reversely implemented} indicates an outcome contrary to the requirement. Score this dimension according to fulfillment as well as implementation quality. Consider the resulting placement together with its spatial relationship. Both should appear natural and coherent.  \\

\midrule
Scoring-band anchors & \textbf{Spatial Layout}---9--10: balanced major-room proportions, no elongated forms, no or only a very short dedicated corridor, appropriate room hierarchy, with no fragmentary space; 7--8: most rooms proportioned appropriately, moderate corridor share, with at most minor local narrowness; 6--7: one or two clear scale problems; 5--6: multiple scale problems, large fragmentary space, excessive corridor share, or reversed master/secondary hierarchy; 0--4: severely unbalanced layout.\newline
\textbf{Physical Compliance}---9--10: $P0=0$, $P1\leq2$; 8--9: $P0=0$, $P1=3$--$6$; 7--8: $P0=0$, $P1\geq7$; 5.5--7: $P0=1$; 3--5.5: $P0=2$; 0--3: $P0\geq3$. P1 issues may gradually lower the score within a band, but must not outweigh P0 defects. \\
\midrule
Hidden design-defect checklist & Living/dining space traversed by circulation and turned into a passage; a severely imbalanced frontage-to-depth ratio creating a strip-like plan; excessive room subdivision causing cramped rooms or fragmentary blocks; a main public space smaller than a secondary room; a bathroom door facing the entry, dining table, or bed; or an open kitchen venting directly toward living/dining space without any partition. \\
\bottomrule
\end{tabularx}
\caption{VLM-HS scoring prompt, Part IV: checklists and scoring anchors.}
\label{tab:vlm_prompt_4}
\end{table*}

\begin{table*}[p]
\centering
\scriptsize
\renewcommand{\arraystretch}{1.08}
\begin{tabularx}{\textwidth}{@{}>{\raggedright\arraybackslash}p{3.4cm}>{\raggedright\arraybackslash}X@{}}
\toprule
\textbf{Prompt component} & \textbf{Instruction content} \\
\midrule
Scoring procedure & \textbf{A. Dimensions 1--5:} First inspect the relevant deduction/bonus checklist together with the hidden-defect checklist. Record only visible hits with their locations. Then assign a score based on the number and severity of hits rather than a vague overall impression: four or more deductions, or one very severe deduction, normally yields 5--6; three deductions yields about 6; two deductions yields 7--8; at most one minor deduction yields about 8; no deduction yields 8--9. Multiple visible bonuses with no deduction may yield 10. Scores of 6 or below must name the specific deduction and location; scores of 8 or above must be supported by no obvious deduction. \newline
For Dimension 5, audit every design-point ID separately. Each item must be labeled \emph{Implemented}, \emph{Partially implemented}, \emph{Not implemented}, or \emph{Reversely implemented}, with image-grounded evidence. Do not merge multiple IDs using statements such as ``most'' or ``all.'' A design point is implemented only when its key action and intended result are both visible; unclear evidence is labeled partially implemented. The dimension score reflects both fulfillment and implementation quality, but is not mechanically fixed by the proportion of item statuses. \newline
\textbf{B. Dimension 6:} Use the P0/P1 fact list as the primary basis. Cross-check obvious inconsistencies against the image, do not fabricate additional issues, and assign the score using the P0/P1 anchors. A true $P0$ must substantially lower the score: $P0=0$ normally implies at least 7, $P0=1$ cannot exceed 7, and $P0\geq2$ cannot exceed 5.5. \\
\midrule
Output format & Output only one strict JSON object, with no draft reasoning or other text. The \texttt{design\_point\_checks} array is mandatory: every supplied design-point ID must occur exactly once; output an empty array if no design points are supplied. Reasons must be concise, image-grounded, and consistent with the item-level audit. Each reason states at most two deductions or bonuses with locations; vague boilerplate is prohibited. If a dimension cannot be judged from the rendering, use score $-1$ and explain why; score 0 is reserved for an extremely poor but judgeable plan. Do not output an overall score. The JSON has the form:\newline
\texttt{\{"design\_point\_checks": [\{"id": "L2-1", "status": "Implemented", "evidence": "..."\}], "dimensions": [\{"dimension": "Functional Zoning", "score": <0--10>, "reason": "..."\}, \{"dimension": "Spatial Layout", ...\}, \{"dimension": "Circulation Quality", ...\}, \{"dimension": "Amenity Configuration", ...\}, \{"dimension": "Design-Point Implementation", ...\}, \{"dimension": "Physical Compliance", ...\}]\}} \\
\bottomrule
\end{tabularx}
\caption{VLM-HS scoring prompt, Part V: scoring procedure and output format.}
\label{tab:vlm_prompt_5}
\end{table*}

\FloatBarrier


\setcounter{secnumdepth}{0}
\end{document}